\documentclass[sigconf]{acmart}
\usepackage{subfig}
\def\BibTeX{{\rm B\kern-.05em{\sc i\kern-.025em b}\kern-.08emT\kern-.1667em\lower.7ex\hbox{E}\kern-.125emX}}
    
\copyrightyear{2019}
\acmYear{2019}
\setcopyright{acmlicensed}
\acmConference[KDD 2019 Workshop]{AI for fashion
The fourth international workshop on fashion and KDD}{August 2019}{Anchorage, Alaska - USA}
\acmBooktitle{KDD 2019 Workshop: AI for fashion
The fourth international workshop on fashion and KDD, August 2019, Anchorage, Alaska - USA}
\acmPrice{15.00}
\acmDOI{10.1145/1122445.1122456}
\acmISBN{978-1-4503-9999-9/18/06} 

\begin{document}

%
\title{Progressive Fashion Attribute Extraction}

%
\author{Sandeep Singh Adhikari}
\email{sandeep.sadhikari@myntra.com}
\affiliation{%
  \institution{Myntra Designs Pvt Ltd}
  \city{Bangalore}
}

\author{Sukhneer Singh}
\email{sukhneer.singh@myntra.com}
\affiliation{%
  \institution{Myntra Designs Pvt Ltd}
  \city{Bangalore}
}

\author{Anoop Rajagopal}
\email{anoop.kr@myntra.com}
\affiliation{%
  \institution{Myntra Designs Pvt Ltd}
  \city{Bangalore}
}
\author{Aruna Rajan}
\email{aruna.rajan@myntra.com}
\affiliation{%
  \institution{Myntra Designs Pvt Ltd}
  \city{Bangalore}
}
%
\renewcommand{\shortauthors}{Adhikari, et al.}

%
\begin{abstract}
Extracting fashion attributes from images of people wearing clothing/fashion accessories is a very hard  multi-class classification problem. Most often, even catalogues of fashion do not have all the fine-grained attributes tagged due to prohibitive cost of annotation. Using images of fashion articles, running multi-class attribute extraction with a single model for all kinds of attributes (neck design detailing, sleeves detailing, etc) requires classifiers that are robust to missing and ambiguously labelled data. In this work, we propose a progressive training approach for such multi-class classification, where weights learnt from an attribute are fine tuned for another attribute of the same fashion article (say, dresses). We branch networks for each attributes from a base network progressively during training.While it may have many labels, an image doesn't need to have all possible labels for fashion articles present in it. We also compare our approach to multi-label classification, and demonstrate improvements over overall classification accuracies using our approach. 
\end{abstract}

%
%

%

%

%
\maketitle

\section{Introduction}
Fashion attribute classification is an important problem for understanding what people wear with internet data, as well as for cataloguing and a several other applications in fashion e-commerce. The challenge in building classifiers for fashion attributes from images stems from the fact that most images on the internet even on fashion catalogues, are poorly tagged for details. These are left to the viewer/consumer to parse, and rarely outlined in detail given the prohibitive cost of annotation. Given this, it is very important to build a classifier that can automatically recognise relevant fashion attributes from an image, and this is very helpful in building both a taxonomy for fashion attributes as well as understanding any trends or fashion events in terms of these fine grained attributes. 

\begin{figure*}[ht]
\includegraphics[height=3in, width=1.8\columnwidth]{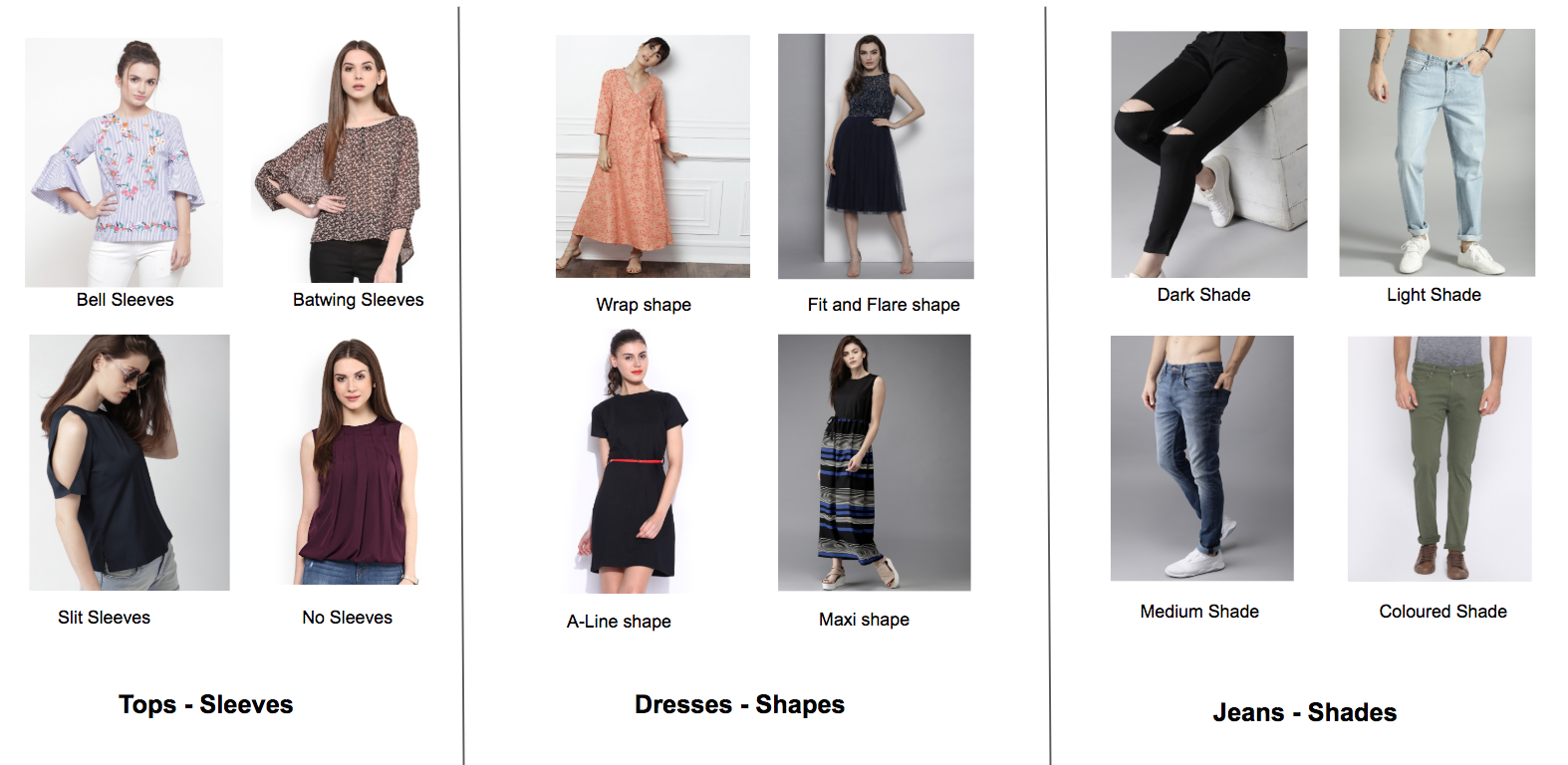}
\caption{Fashion articles and their attributes.\label{all_articles}}
\end{figure*}

\begin{figure*}[!ht] 
\centering
  \subfloat[Top]{%
    \includegraphics[height=2in, width=0.8\columnwidth]{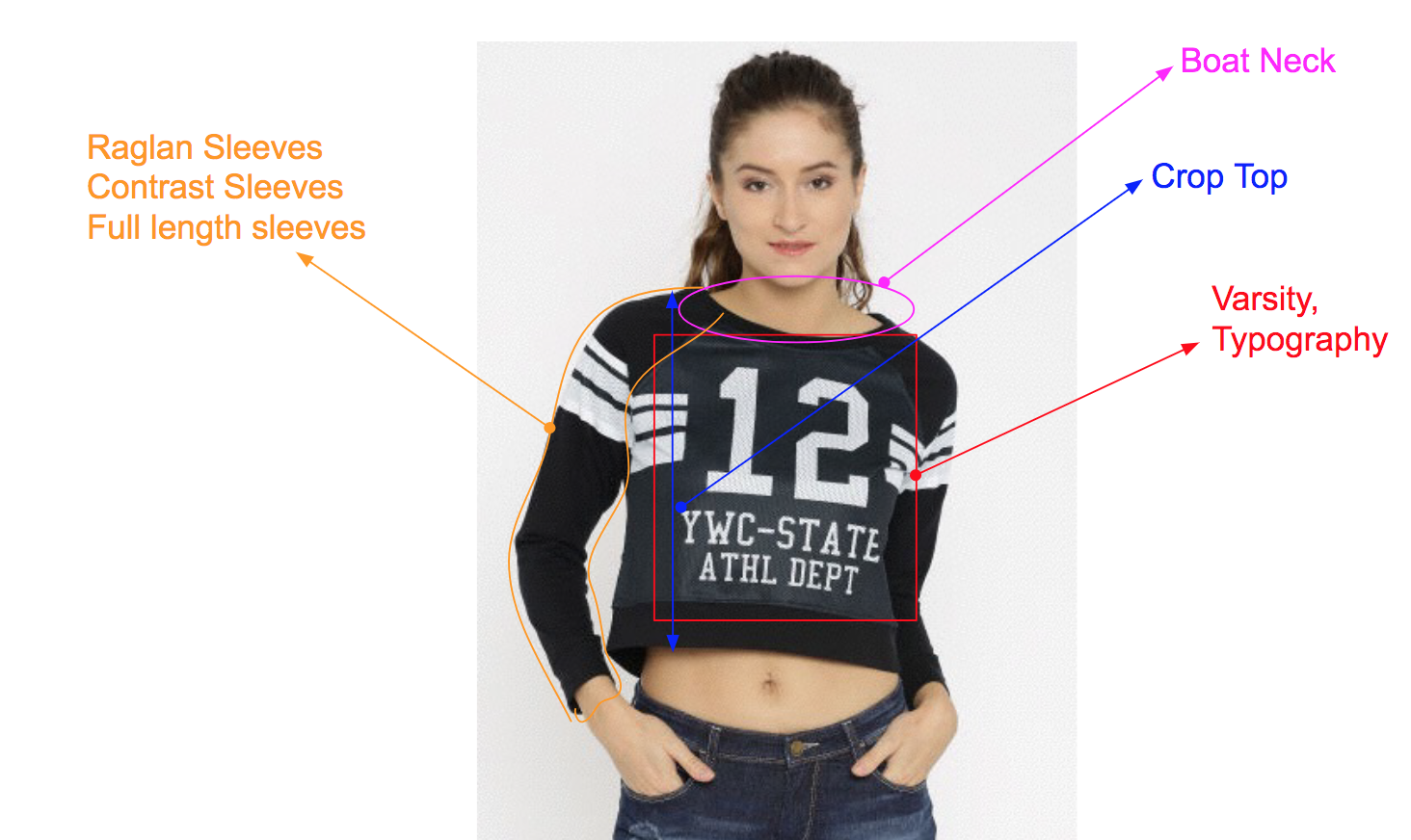} 
  } 
  \subfloat[Dress]{%
    \includegraphics[height=2in, width=0.6\columnwidth]{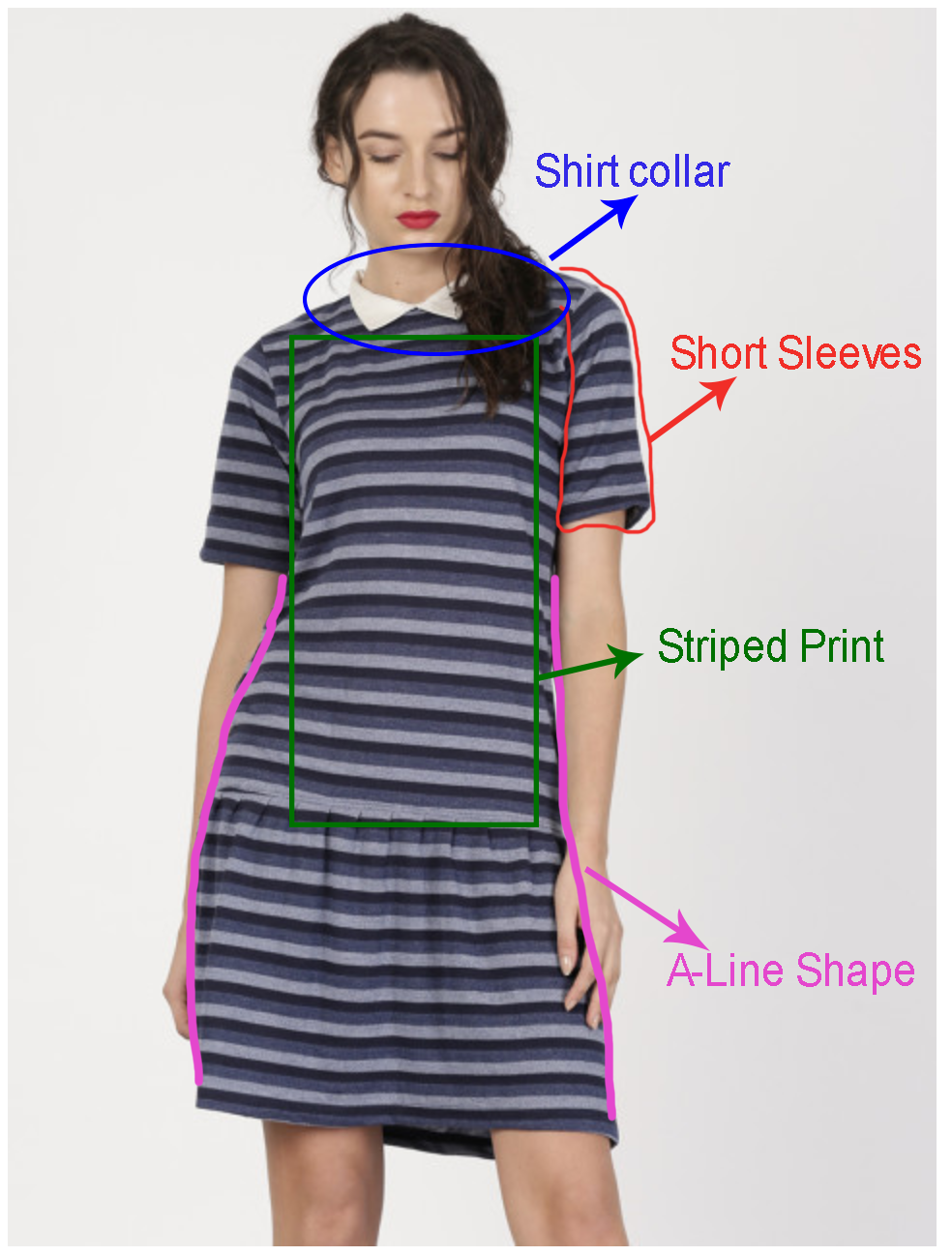} 
  } 
  \caption{Tops and Dress articles and their corresponding attributes}
  \label{tops_dress} 
\end{figure*}

Different attributes define different fashion articles. For instance, the shape of a dress is very important to its styling, where as the neck and sleeve styling stand out in a top(see Figure~\ref{all_articles}). Example of a fashion articles from tops and dresses and their corresponding attributes are shown in Figure~\ref{tops_dress}. Even within a given attribute such a sleeve type, there are distinctive varieties that stand out because of their stylistic details in the design. This makes the problem both challenging and interesting to study.

Given the sparsity of labelling and annotation that cover various stylistic elements or attributes of a fashion article, we make the following contributions to the extraction of fashion attributes from images: 
(i) We detail an approach to progressively train classifiers to identify fashion attributes where certain attributes can be missing or incompletely labelled. This is a more practical approach than multilabel classifiers as labelled data is always sparse and scarce in the world of fashion. 
(ii) Our model has a base network which is common for all the attributes and a separate branched network for each attribute. Since we train our model in progressive manner, an image doesn't need to have labels for all the attributes. This is different from the multi-label classification approach where an image must have all the labels present. We show results comparing our approach to individual attributes classifiers and multilabel classifier. Our method performs slightly better than individual classifier models and is slightly better than the multilabel classifier approach.

\section{Related work}
Deep convolutional neural networks are now state-of-art on performing many visual tasks, specially because of large-scale annotated image datasets such as ImageNet~\cite{deng2009imagenet}. Residual Networks (ResNet)~\cite{he2016deep} is one of the state-of-art deep learning model for visual classification. Fashion attribute extraction is one of the challenging task as there is a huge variety in various style images and also there might exist a huge correlation among labels.
A unified model for multilabel classification is presented in~\cite{ferreira2018unified}. Here they try solve the classification in hierarchical way by a common deep learning architecture predicting category, sub category and attributes of the dresses. In~\cite{inoue2017multi} multilabel fashion attribute prediction is proposed under label noise condition where a label cleaning network along with a classification network is used which works on a set of $66$ weakly labelled fashion attributes. In~\cite{dong2017multi} authors propose a model transfer learning from well-controlled shop clothing images collected from web retailers to in-the-wild images from the street. In~\cite{liu2016deepfashion} a dataset of huge collection of attributes is provided across various articles. A weighted cross entropy is utilized to predict different attributes in an image. However, these works require that the image is provided with all its multi labels through some human curation to work well which is an expensive procedure.

In our work, we propose to circumvent the requirement of all labels to be annotated for a multi label classification by adopting a progressive training procedure using multiple multi-class classification. Thus we can utilize all data for a given attribute (say neck type) even if other attributes of it is not annotated (say print). We propose a progressive architecutre with training inspired from Progressive gans~\cite{karras2017progressive} for attribute extraction.

\section{Progressive training}
Our goal is to predict the different attributes that are present given an input image of a fashion article (worn by a model). Attributes we use in our training are curated by merchandisers and category managers, and are as mentioned in Table~\ref{table:attributes}

\begin{table*}
\begin{center}
\caption{Different articles and their corresponding attributes. The number of values in each attribute is shown in brackets.}
\label{table:attributes}
{\begin{tabular}{ll}
\hline\noalign{\smallskip}
Article & Attributes\\
\noalign{\smallskip}
\hline
\noalign{\smallskip}
Dresses  & Shape(15), Length(4), Hemline(7), Sleeve Style(14),Pattern(19),\\
             & Sleeve length(4), Neck(16)\\
Tops & Sleeve Style(18), Sleeve Length(5), Pattern(18),  Neck(14)  \\
Jeans & Fade(3), Shade(4), Distress (5)\\
\hline
\end{tabular}}
\end{center}
\end{table*}

\subsection{Individual models for attributes}
For the well defined task of visual classification for attributes, we first experiment with Resnet-34 architecture for each of the individual attributes mentioned in Table~\ref{table:attributes}. We use this as baseline individual models. We used pretrained ImageNet weights for initialization purpose and trained all the layers using differential learning rates.

 \begin{figure*}[ht]
\includegraphics[height=1.3in, width=1.5\columnwidth]{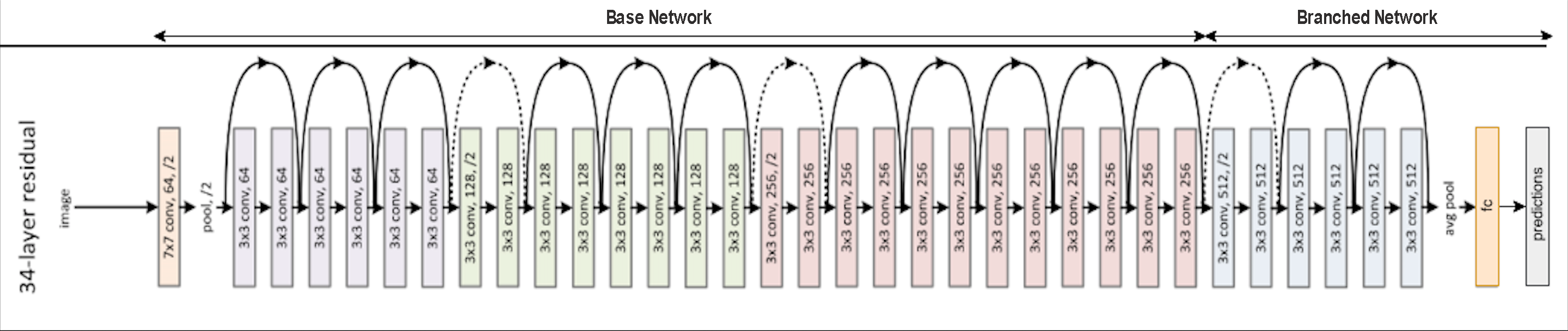}
\caption{Resnet-34 architecture.\label{fig:resnet}}
\end{figure*}

 \begin{figure*}[ht]
 \centering
\includegraphics[height=2in, width=1.2\columnwidth]{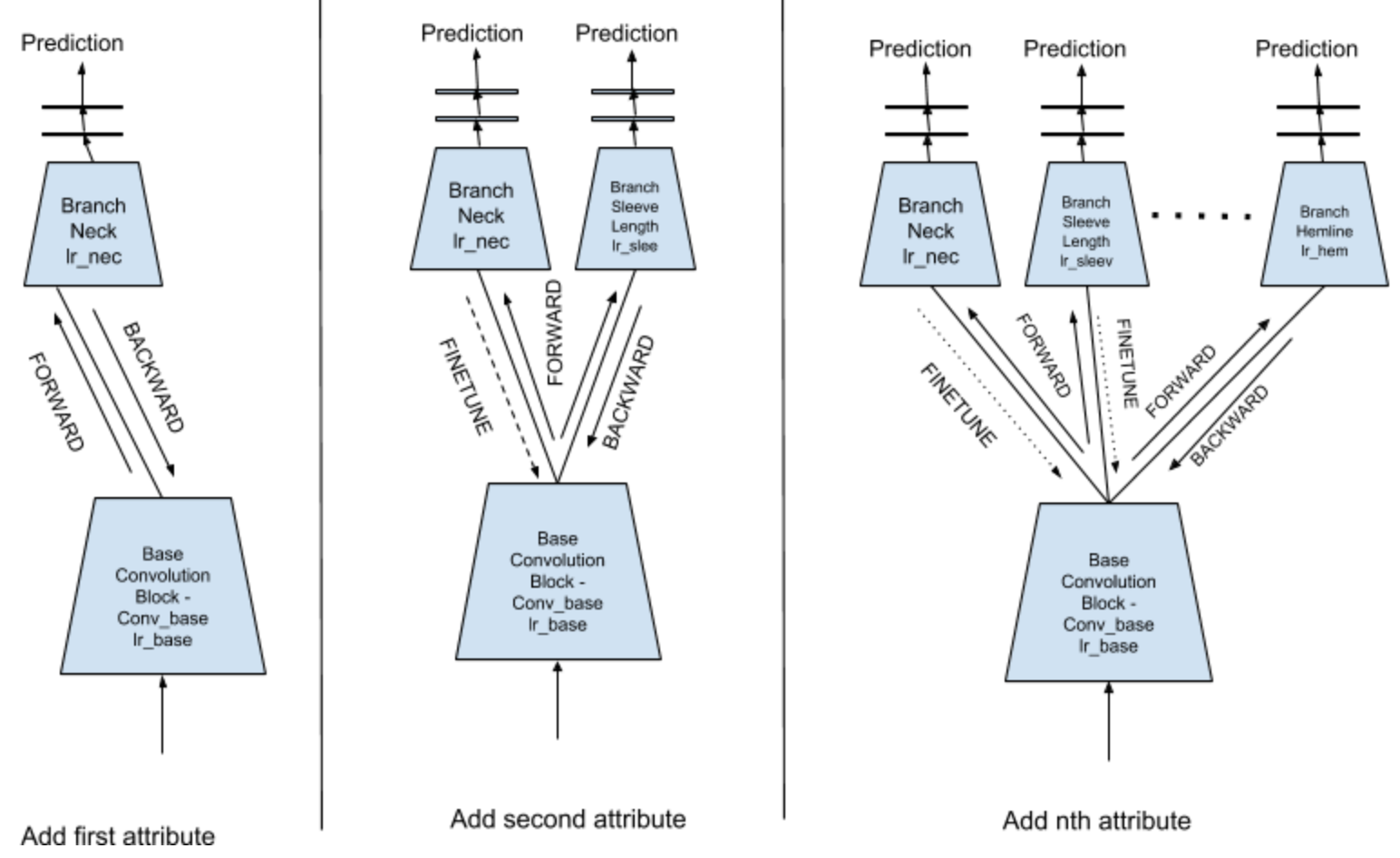}
\caption{Our progressive training approach.\label{fig:architecture}}
\end{figure*}
\subsection{Progressive approach for all attributes}
\label{training}
We divide the ResNet-34 architecture into two parts - Base Network (B) and Branch Network as shown in Figure~\ref{fig:resnet}. We add this Branch Network progressively during the training process for each attribute($N_1, N_2,\dots, N_n$ where $n$ is number of attributes). We use the softmax classifier in each of the Branch networks.

We used pre-trained weights of ResNet-34 model (trained on ImageNet[1] datasets) for initialization, and then implement the following training procedure:
\begin{itemize}
\item Base network (B) and each of the Branch network will have pre-trained weights of Resnet-34.
\item Start the training with data of an attribute $A_1$ and update the weights of Base network (B) and Branch network ($N_1$) based on cross entropy loss of attribute $A_1$ for an epoch. We trained all the layers using differential learning rates and kept the learning rate quite low in the initial Base layers and kept on increasing it till the last layer of Branch network.
\item Add Branch network for attribute $A_2$ and train only this Branch and Base network end-end. 
\item Now fine tune the entire network (with a lower learning rate) including the previously added Branches.
\item Repeat the above process adding one Branch network for each attribute at a time till $n$ attributes are added.
\item Finally once all attributes are added and trained, freeze the Base network and fine tune the Branch networks one at a time. (Since parameters of each branch are independent of each other joint optimization can be performed by combining all the loss functions instead of one at a time.)
\end{itemize}

The schematic of our approach is shown in Figure~\ref{fig:architecture}. Thus essentially in an overall epoch, the base network weights get updated by all the attributes data and the branch network weights which are specific to an attribute get updated by that particular attribute. As such it is possible that Branch networks will localize the region where the attribute is present. For instance, if we have to do attribute extraction for neck, the particular branch network could possibly give better activation for region in the image where neck is present. In our experiments, we did find that adding branch networks for each attribute improved the overall accuracy. Note that we can always add a new branch network whenever we see a presence of new attribute. Since base network is trained with all other attributes, we noticed that with such a base initialization our training time also reduced.  

\subsection*{Multi-Label approach}

In this Multi-label classification approach for attribute extraction we one-hot encode all the classes for all the attributes for a given article. We take the pre-trained weights of ResNet-34 architecture and add one more fully connected layer having m nodes, where m is the total number of one-hot encoded classes of all the attributes. We used sigmoid activation followed by multi-label soft margin loss for training that optimizes a multi-label one-versus-all loss based on max-entropy.

\section{Experiments}

We used image datasets of three fashion articles - Dresses, Tops and Jeans. We separately trained models for each of the article types. These images are obtained from Fashion e-commerce sites like Amazon, Asos, Flipkart etc. All the fashion attributes for each image was annotated by humans who are experts in fashion design. Our dataset consisted of $100K$ Tops images, $60K$ Jeans and $60K$ dresses. In general only about $40-50\%$ of the total data had all the attributes tagged. Also if we just look at individual attributes we easily can obtain about nearly $80\%$ of the total data for every attribute during training. However, for evaluation purposes, for multi-label classification training we used $75\%$ of the total data by having the rest of attributes manually tagged for all the articles. We used $80-20$ train-test split in our experiments.

\subsubsection{Evaluation:}
\begin{figure*}[!ht] 
  \subfloat[Sleeve Style (Tops)]{%
    \includegraphics[height=1in,width=0.3\textwidth]{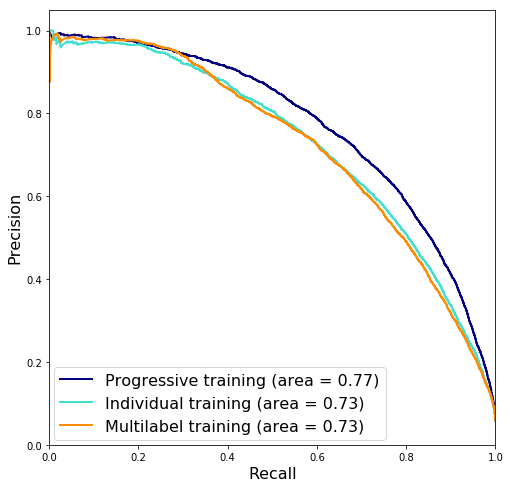} 
  } 
  \hfill 
  \subfloat[Neck(Dresses)]{%
    \includegraphics[height=1in,width=0.3\textwidth]{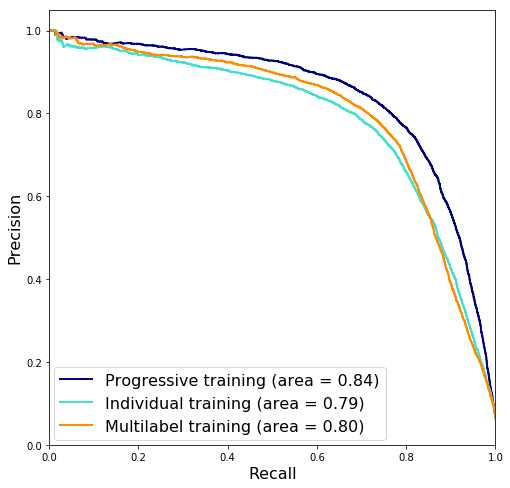} 
  } 
  \hfill 
  \subfloat[Fade(Jeans)]{%
    \includegraphics[height=1in,width=0.3\textwidth]{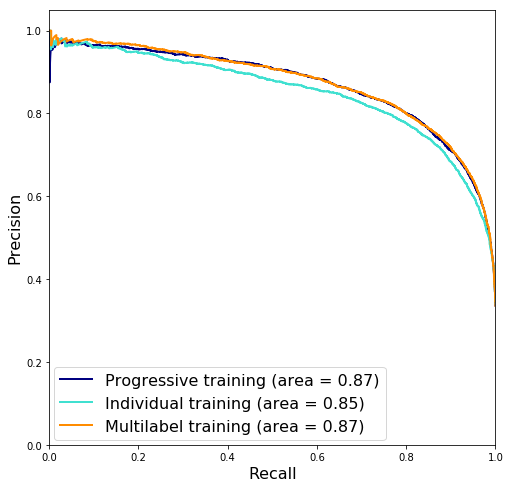} 
  }
  \caption{Precision-Recall Curve}
  \label{fig:PR} 
\end{figure*}

We used multi-class classification accuracy as our evaluation metric for each of the fashion article. We compare our Progressive approach with individual models for each attribute and multilabel model with all attributes for a given article. Tables~\ref{tab:tops}, \ref{tab:jeans} and \ref{tab:dresses} provide classification accuracies for each of the 3 articles. We see that our progressive approach performs better than the individual and multilabel approaches.  
In Figure~\ref{fig:PR} we show precision-recall curves  for Sleeve Style (Tops), Neck(Dresses) and Fade(Jeans) averaged over attribute values respectively. We observe that our approach has better performance overall than the baseline approaches. 
For jeans article we see that our approach and multi-label approach perform almost equally. However, it is noticed that attributes in jeans have less complexity than compared to dresses or tops.

\begin{table*}
\begin{center}
\caption{Accuracies for Tops}
\label{tab:tops}
\scalebox{1}{\begin{tabular}{cccccc}
\hline\noalign{\smallskip}
Model & Sleeve Style & Sleeve Length & Pattern & Neck & Overall\\
\noalign{\smallskip}
\hline
\noalign{\smallskip}
Individual & 65.78 & 87.00 & 67.74 & 70.87 & 72.85\\
Multi-Label & 65.96 & 88.65 & 64.80 & 70.46 & 72.47\\
\textbf{Progressive}  & 69.69 & 88.98 & 68.18 & 75.41 & \textbf{75.57}\\
\hline
\end{tabular}}
\end{center}
\end{table*}

\begin{table*}[ht]
\begin{center}
\caption{Accuracies for Jeans}
\label{tab:jeans}
\scalebox{1}{\begin{tabular}{ccccc}
\hline\noalign{\smallskip}
\small{Model} & \small{Distress} & \small{Fade} & \small{Shade}  & \small{Overall}\\
\noalign{\smallskip}
\hline
\noalign{\smallskip}
Individual & 82.60 & 76.25 & 78.58 & 79.14 \\
Multi-Label & 84.88 & 80.23 & 77.25 & \textbf{80.79}\\
\textbf{Progressive}  & 84.38 & 79.68 & 77.00 & 80.35 \\
\hline
\end{tabular}}
\end{center}
\end{table*}

\begin{table*}
\begin{center}
\caption{Accuracies for Dresses}
\label{tab:dresses}
{\begin{tabular}{lllllllll}
\hline\noalign{\smallskip}
Model & Sleeve & Length  & Sleeve & Neck & Hemline & Shape & Pattern & Overall\\
          & Length &             & Style  &          &             &           &             & \\
\noalign{\smallskip}
\hline
\noalign{\smallskip}

Individual & 89.2 & 79.03 & 75.59 & 74.33 & 71.37 & 64.54 & 66.66 & 71.92\\
Multi-Label & 89.57 & 80.39	& 76.01	& 75.54 &  72.20 & 66.69 & 64.74 & 75.02\\
\textbf{Progressive} & 91.11 & 79.06 & 78.24 & 78.12 & 71.88& 68.07 & 68.34 & \textbf{76.40}\\

\hline
\end{tabular}}
\end{center}
\end{table*}
\setlength{\tabcolsep}{1.4pt}

\section{Discussion}
In this section we discuss some of the benefits of taking this approach for production. The Progressive Learning Model offers many advantages in deployment to production environment compared to Individual models. Though the model complexity is slightly high in terms of number of branches, it has the benefit of being modular. That is, it is very easy to add a new attribute by just adding a branch. Note that we also retrained our system by different reordering of the attributes and the results did not vary much (Eg. for jeans, we trained distress $>$ fade $>$ shade, fade $>$ shade $>$ distress etc..in all $3$ factorial ways). This probably indicates that we can add a new attribute at any point of time without disturbing the performance of the other attributes.

To get all $n$ attributes for a dress we need to have $n$ different models to be deployed in the production if we were to train individual models. In our corpus we have nearly $300$ different types of fashion articles. Even if we consider approximately around $4$ attributes to each on an average we are looking at around $1200$ models. Maintaining these many models and their different versions becomes cumbersome. Whereas in our case we just need to maintain $300$ models. Also coming to hosting the service, since we have only one model for a given article, we can host only one service end points per article. Also given an input image, it need not be stored in memory as in individual models to run on each attributes. 

Moreover, the size of Progressive Learning Model is also lesser than the combined size of all individual models since the base network is common for all the attributes. This helps in scaling up as we can fit more models in a given system with limited RAM. The size taken up is shown in Table~\ref{tab:size}

\begin{table}
\begin{center}
\caption{Model Size on disk for all attributes}
\label{tab:size}
\scalebox{1}{\begin{tabular}{c|c|c}
\hline\noalign{\smallskip}
Article Type & Individual models & Progressive models \\
\noalign{\smallskip}
\hline
\noalign{\smallskip}
Dresses(7 attributes) & 1120MB & 450MB\\
Tops(5 attributes) & 800MB & 350MB\\
Jeans(3 attributes)  & 480MB & 230MB\\
\hline
\end{tabular}}
\end{center}
\end{table}

\subsection{Model Training and Inference Time}

Our Progressive Learning Model takes almost $\sim{50}\%$ lesser time to get trained than combined time of all individual models. This is because after certain overall epoch number, only the branched network weights get updated and base network weights get freezed.   

The Inference time is also get reduced by almost $\sim{40}\%$ because forward pass is required only once till the end of the base network. The time taken up is summarized in Table~\ref{tab:time}. We clearly observe that our way of architecture has a distinct advantage of deploying individual models both in cpu or gpu. Also scaling becomes easier on cpu if parallelized on the multiple systems.

\begin{table*}
\begin{center}
\caption{Model inference time for all attributes}
\label{tab:time}
\scalebox{1}{\begin{tabular}{c|c|c|c|c}
\hline\noalign{\smallskip}
Article Type & Individual models(gpu) & Progressive models(gpu) & Individual models(cpu) & Progressive models(cpu) \\
\noalign{\smallskip}
\hline
\noalign{\smallskip}
Dresses(7 attributes) & 0.28s & 0.12s & 1.4s & 0.55s\\
Tops(5 attributes) & 0.2s & 0.1s & 1.1s & 0.46s\\
Jeans(3 attributes)  &0.14s & 0.08s & 0.7s & 0.34s\\
\hline
\end{tabular}}
\end{center}
\end{table*}

\subsection{Production Setup}

\begin{figure*}[ht]
\includegraphics[height=3in, width=1.5\columnwidth]{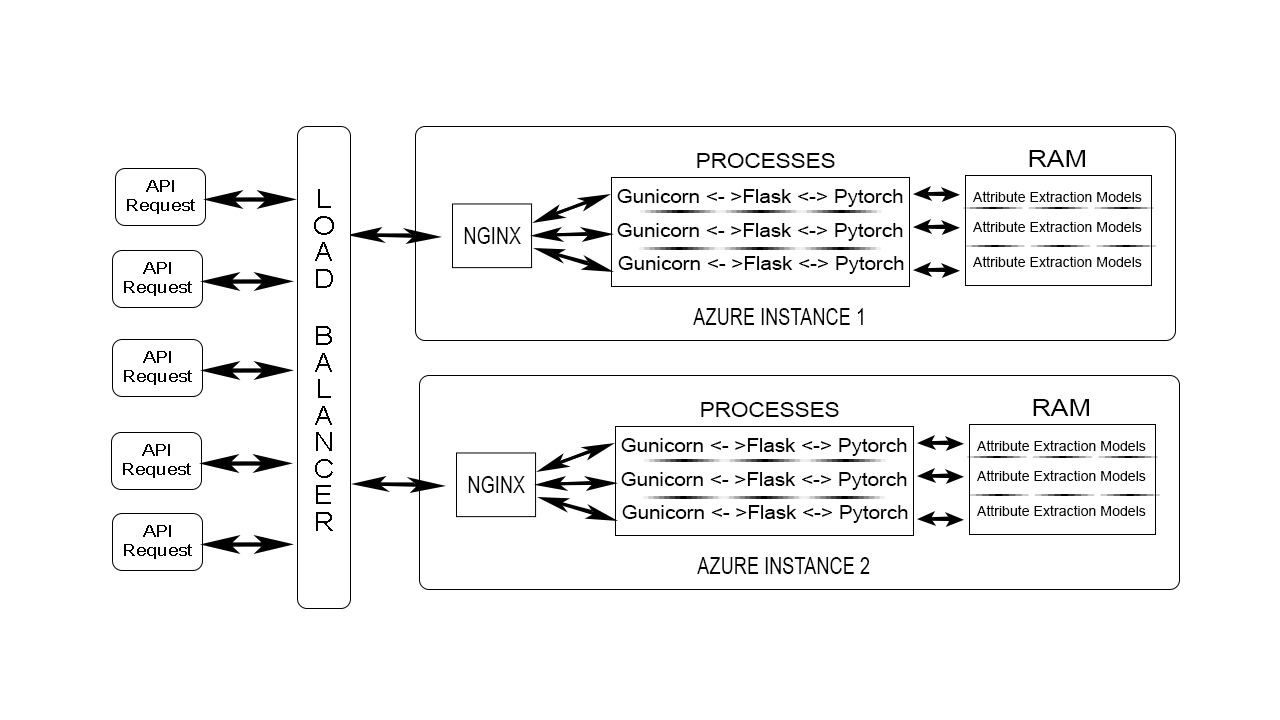}
\caption{Production schema.\label{fig:prod}}
\end{figure*}

We scaled our models using multiple azure instances and then running multiple instances of models in each of these instances. All of these is handled using Flask framework with gunicorn and nginx servers. The schema is presented in Figure~\ref{fig:prod}

Details of each component in productions:
\begin{itemize}
    \item Attribute Extraction Models:
Three Instances of each model per article are loaded in RAM, by separate process of python. These models are handled by pytorch in which whole code is written.

     \item Flask:
	Flask acts as a handler for pytorch's code. Three instances of flask take care of models separately.
	
	\item Gunicorn:
	Gunicorn is an interface between flask and NGinx server. This runs required handler of flask which internally executes the code.
	
	\item Nginx:
	Nginx servers act as an internal load balancer, which assigns request received to any of three Gunicorn instances. This takes cares of queuing and slow requests.
	
	\item Load Balancer:
	Provided by Azure, this load balancer decides on which instance to send request on, considering load on all instances.
\end{itemize}

\subsection{Zero downtime for requests}
This parallel deployment of model as separate instances enables us to update model files on the go. Usually any particular instance of azure is brought down to make changes, and meanwhile other serves the request. New updates are then checked with directed requests(using IP address, without load balancer), if everything is fine updated instance is then added to load balancer.

\section{Conclusions}

In this paper we have proposed a method to progressively grow a resnet architecture to perform multilabel classification, and compared the performances of our progressive Learning Model with Multi-label classification and Individual models trained for attribute extraction. We show that our Progressive Learning Model always performs better than individual models and is par with the multi-label classification model. But its main advantage is that unlike multi-label classification model,  it can leverage all the images data even if the image does not have label for all the attributes, thus being more robust to errors and missing data, as well as needing a lot less training data. We also provide a production set up for our models and correspondingly scaling them. We also provide various performance measures of our models in deployment. As a future work, we can automate this whole process of adding Branch networks whenever a new attribute is added either in the catalog or in the test description provided in an e-commerce setting.

%
\bibliographystyle{ACM-Reference-Format}
\bibliography{acmart}

%

\end{document}